\documentclass[journal]{IEEEtran}
\usepackage{caption}
\usepackage{graphicx}
\usepackage{epstopdf}
\usepackage{dcolumn}
\usepackage{bm}
\usepackage{float}
\usepackage{color}
\usepackage{subfigure}
\usepackage[utf8]{inputenc}
\usepackage[T1]{fontenc}
\usepackage{mathptmx}
\usepackage{booktabs} 

\usepackage{algorithm}
\usepackage{algorithmic}

\usepackage{indentfirst}

\usepackage{mathptmx}

\usepackage[font=small,labelfont=bf,labelsep=period]{caption}

\ifCLASSINFOpdf
\else
\fi
\begin{document}
%
\title{DanHAR: Dual Attention Network For Multimodal Human Activity Recognition Using Wearable Sensors}
%
%
%

\author{Wenbin Gao, Lei Zhang, Qi Teng, Jun He, Hao Wu

\thanks{This work was supported  in part by the National Science Foundation of China under Grant 61971228 and the Industry-Academia Cooperation Innovation Fund Projection of Jiangsu Province under Grant BY2016001-02, and in part by the Natural Science Foundation of Jiangsu Province under grant BK20191371.\textit{ (Corresponding author: Lei Zhang.)}}
\thanks{Wenbin Gao, Lei Zhang and Qi Teng are with School of Electrical and Automation Engineering, Nanjing Normal University, Nanjing, 210023, China (e-mail: leizhang@njnu.edu.cn).}
\thanks{Jun He is with School of Electronic and Information Engineering, Nanjing
	University of Information Science and Technology, Nanjing 210044, China.}
\thanks{Hao Wu is with School of Information Science and Engineering, Yunnan University, Kunming 650091, China.}
\thanks{This work has been submitted to the ELSEVIER for possible publication. Copyright may be transferred without notice, after which this version may no longer be accessible.}

%

}


%



\maketitle

\begin{abstract}
Human activity recognition (HAR) in ubiquitous computing has been beginning to incorporate attention into the context of deep neural networks (DNNs), in which the rich sensing data from multimodal sensors such as accelerometer and gyroscope is used to infer human activities. Recently, two attention methods are proposed via combining with Gated Recurrent Units (GRU) and Long Short-Term Memory (LSTM) network, which can capture the dependencies of sensing signals in both spatial and temporal domains simultaneously. However, recurrent networks often have a weak feature representing power compared with convolutional neural networks (CNNs). On the other hand, two attention, i.e., hard attention and soft attention, are applied in temporal domains via combining with CNN, which pay more attention to the target activity from a long sequence. However, they can only tell where to focus and miss channel information, which plays an important role in deciding what to focus. As a result, they fail to address the spatial-temporal dependencies of multimodal sensing signals, compared with attention-based GRU or LSTM. In the paper, we propose a novel dual attention method called DanHAR, which introduces the framework of blending channel attention and temporal attention on a CNN, demonstrating superiority in improving the comprehensibility for multimodal HAR. Extensive experiments on four public HAR datasets and weakly labeled dataset show that DanHAR achieves state-of-the-art performance with negligible overhead of parameters. Furthermore, visualizing analysis is provided to show that our attention can amplifies more important sensor modalities and timesteps during classification, which agrees well with human common intuition.\\
\end{abstract}


%
\IEEEpeerreviewmaketitle

\section{Introduction}
%
%
%
%
\IEEEPARstart{T}{he} last few years have seen the success of ubiquitous sensing, which aims to extract knowledge from the data acquired by pervasive sensors \cite{avci2010activity}. Human activity recognition (HAR) using wearable sensors has become a very active research within the field, especially for various real-world applications such as sports, interactive gaming, health care, and monitoring systems for general purposes. In essence, multimodal HAR can be seen as a classic multivariate time series classification problem, which uses sliding time window to segment time series sensor signals and extracts discriminative features \cite{banos2014window}. Each time window can further be recognized by utilizing conventional machine learning methods such as decision tree, k-nearest neighbors and support vector machine \cite{bulling2014tutorial}. Though these shallow learning methods have made considerable achievements on inferring activity details, they heavily rely on hand-crafted feature extraction requiring the knowledge of experts \cite{cheng2020real}, which are task or application dependent and cannot be transferred to similar activity recognition  tasks. Furthermore, the shallow learning is hard to represent the salient characteristics of complex activities, and usually involves time-consuming process to select optimal features \cite{tang2020efficient}. To solve above problems, the studies that delve into automatic feature extraction with less human efforts have become a critical research area. Multimodal HAR research is undergoing a transition from shallow learning to deep learning \cite{alsheikh2016deep}.\\
\indent Recently, deep learning has represented an important research trend in sensor based HAR, in which different layers are stacked to form deep neural networks (DNNs) that provide better system performance and remove the need for hand-crafted features \cite{wang2019deep}. In particular, convolutional neural networks (CNNs) have significantly pushed the performance of HAR due to their rich representation power. With increasing model capacity for rich representation, DNNs will have higher performance, but inevitably lead to more requirements for strictly labelled data. One challenge for deep HAR recognition is the collection of annotated or “ground truth labeled” training data \cite{he2018weakly}. Ground truth annotation is an expensive and tedious task, in which the annotator has to perform annotation via skimming through raw sensor data and manually label all activity instances. However, the time series data recorded from multimodal embedded sensors such as accelerometer or gyroscope is far more difficult to interpret than data from other sensors modalities, such as cameras. It requires laborious human efforts to accurately segment and label a target activity from a long sequence of time series sensor data. On the whole, although these DNN models can automatically extract appropriate features for classification, they still require strict annotation to label their ground-truth, which would cost much more human efforts to create a perfect training dataset for HAR in supervised learning scenario.\\
\indent Intuitively, it is much easier for the human beings who are recording sensor data to identify whether an interesting activity takes place in a long sequence. If we can infer activity kinds from the coarse-grained labels and determine specific locations of every labeled target activity with only knowledge of which kinds of activities contained in the long sequence, it will greatly reduce the burden of manual labeling \cite{wang2020sequential}. Therefore, it deserves further research whether we can directly recognize and locate a target activity from coarsely labeled sensor data. We tackle the above challenges from a different aspect, i.e., attention, which recently has been studied extensively in various research areas such as computer vision \cite{zagoruyko2016paying} and natural language processing (NLP) \cite{zhou2016attention}. Similar to the human’s perception, attention attempts to focus selectively on parts of the target areas to enhance interesting details of the targets while suppressing other irrelevant and potentially confusing information. That is to say, attention can tell where to focus via improving the representation of interests.\\
\indent To our knowledge, attention has seldom been used in HAR scenario. Recently, two attention mechanisms are proposed via combining with a Gated Recurrent Units (GRU) network \cite{ma2019attnsense} and a Long Short-Term Memory (LSTM) network \cite{zeng2018understanding} respectively, which can capture the dependencies of sensing signals in both spatial and temporal domains simultaneously. However, recurrent networks often have a weak feature representing power compared with CNN. In our previous work, two attention mechanisms, i.e., hard attention \cite{he2018weakly} and soft attention \cite{wang2019attention}, are introduced via combining with CNN, which can pay more attention to the target activity from a long sequence. As a result, they can only tell where to focus and miss the channel information, which plays an important role in deciding what to focus. In other words, the two existing methods fail to address the spatial-temporal dependencies of multimodal sensing signals, compared with attention-based GRU or LSTM. In computer vision field, convolution operations usually extract features via blending cross-channel and spatial information together. Recently a series of researches have been introduced to incorporate channel attention into convolution blocks \cite{hu2018squeeze,gao2019global,woo2018cbam,roy2018recalibrating,cao2019gcnet,fu2019dual}, showing great potential in performance improvement.\\ 
\indent The channel attention has never been considered in HAR. Inspired by the idea, we aim to increase representation power and comprehensibility by incorporating the channel attention for multimodal HAR. In the paper, we for the first time propose a novel dual attention method called DanHAR in multimodal HAR scenario, which blends channel and temporal attention on a CNN model. To increase feature extraction capacity, residual network is also introduced as our backbone. We sequentially infer channel and temporal attention maps, which enable us to learn what and where in both spatial and temporal domains simultaneously. The proposed attention can capture channel features and temporal dependencies of multimodal time series sensor signals, which amplifies the more important sensor modalites and timesteps during classification. Extensive experiments are conducted to evaluate DanHAR on four public benchmark HAR datasets consisting of WISDM dataset \cite{kwapisz2011activity}, PAMAP2 dataset \cite{reiss2012introducing}, UNIMIB SHAR dataset \cite{micucci2017unimib} and OPPORTUNITY dataset \cite{chavarriaga2013opportunity}, as well as the weakly labeled dataset. We show that exploiting both is superior to using one attention alone. Visualizing analysis of attention weights is further provided to explore how attention has focused on multimodal time series sensor signals to improve the model’s comprehensibility. The experimental results indicate the advantage of DanHAR blending channel and temporal attention with regard to typical challenges in multimodal HAR scenarios.\\
\indent Our main contribution is three-fold. Firstly, we for the first time propose an efficient DanHAR method in multimodal HAR scenarios, which can improve representation power of CNN via blending channel and temporal attention together. Secondly, we perform extensive experiments to verify that the attention method achieves state-of-the-art performance with negligible overhead of parameter, across the multiple public benchmark HAR datasets as well as the weakly labeled HAR dataset. Thirdly, various ablation experiments are conducted to validate the effectiveness and efficiency of the attention method. Visualizing analysis of attention weights in both spatial and temporal domains is provided to improve the multimodal sensor signal’s comprehensibility. In addition, the proposed method can effectively aid to the collection of “ground truth labeled” training data.\\
\indent The rest of paper is organized as follows. In Section II, we review the related works of sensor based HAR and attention method. Section III details the proposed attention framework for multimodal HAR. In Section IV, we describe the details of the public HAR dataset and the collected weakly labeled dataset for our study, as well as the details of the experimental design, and give the experimental result comparison and analysis from several aspects. Section V concludes the paper and discusses our future work.

\section{Related Works}
\indent Deep CNNs have been widely used in computer vision community, and many researches are continuously investigated to further improve performance \cite{shin2016deep}. It is well known that attention plays an important part in human perception, in which one does not need to process a whole scene, instead can selectively focus on the most salient parts. Recent advances in a broad range of tasks such as image classification, object detection and semantic segmentation have seen the success of attention \cite{zagoruyko2016paying,zhou2016attention,chen20182}. In particular, incorporation of channel attention into convolution blocks shows great potential in performance improvement, which has attracted a lot of interests from researchers. The squeeze-and-excitation network (SENet) \cite{hu2018squeeze} is one of the representative methods, which brings clear performance gain across various deep CNN architectures via learning channel attention for each convolution block. Inspired by the setting of squeeze and excitation in SENet \cite{hu2018squeeze}, some researchers continue to develop channel-wise attention modules, which can be roughly divided into two directions:(1) improvement of feature aggregation ability; (2) combination of channel and spatial attentions. GSoP \cite{gao2019global} proposes a second-order pooling technique that can achieve better feature aggregation performance. CBAM \cite{woo2018cbam} and scSE \cite{roy2018recalibrating} blend channel attention with spatial attention that use a 2D convolution of kernel size $k\times k$. GC-Net \cite{cao2019gcnet} develops a simplified Non-Local neural network, which is then integrated with SE block, resulting in a lightweight module. Dual Attention Network (DAN) \cite{fu2019dual} simultaneously utilizes Non-Local based channel and spatial attention for semantic segmentation. All above methods have yielded better performance on various computer vision tasks via learning effective channel attention.\\ 
\indent In another line of research, deep learning with automatic feature extraction has become the dominant technique in multimodal HAR, which can automatically learn intricate feature representation from raw sensor readings and discover the best pattern to improve recognition performance. During recent years, there has been an increasing research interest in deep HAR model with attention mechanism. Zeng \emph{et~al.} \cite{zeng2018understanding} utilized LSTM as backbone to incorporate two attention mechanisms for HAR: temporal attention and sensor attention, which can highlight the important part of time series sensor signals, as well as multiple sensor modalities. To maintain the continuity of time series signals, the regularization terms for temporal attention and sensor attention have to be introduced into the LSTM. Ma \emph{et~al.} \cite{ma2019attnsense} proposed an attention method called AttnSense for multimodal sensing HAR, which combines attention mechanism with a GRU network to capture the dependencies of sensing signals in both spatial and temporal domains. However, the aforementioned two attention methods have been mainly applied in recurrent architectures, which has weak feature extraction and classification capacity. Recently, in weakly supervised HAR scenario, He \emph{et~al.} \cite{he2018weakly} proposed a novel CNN using an idea of hard attention, which is non-differentiable and requires more complicated techniques such as reinforcement learning to train. We proposed an end-to-end trainable soft attention CNN via inferring attention maps along temporal dimension, which can focus on the interesting part of the target activity and suppress irrelevant information \cite{wang2019attention}. To the best of our knowledge, the channel-wise attention idea has never been applied in research area of multimodal HAR. Different from previous works, in the paper, we for the first time propose to deal with multimodal HAR tasks via emphasizing meaningful features along two principal dimensions: channel and temporal axes. 

\begin{figure}[htbp]
	\hspace{-0.2cm}
	\includegraphics[height=6cm,width=9cm]{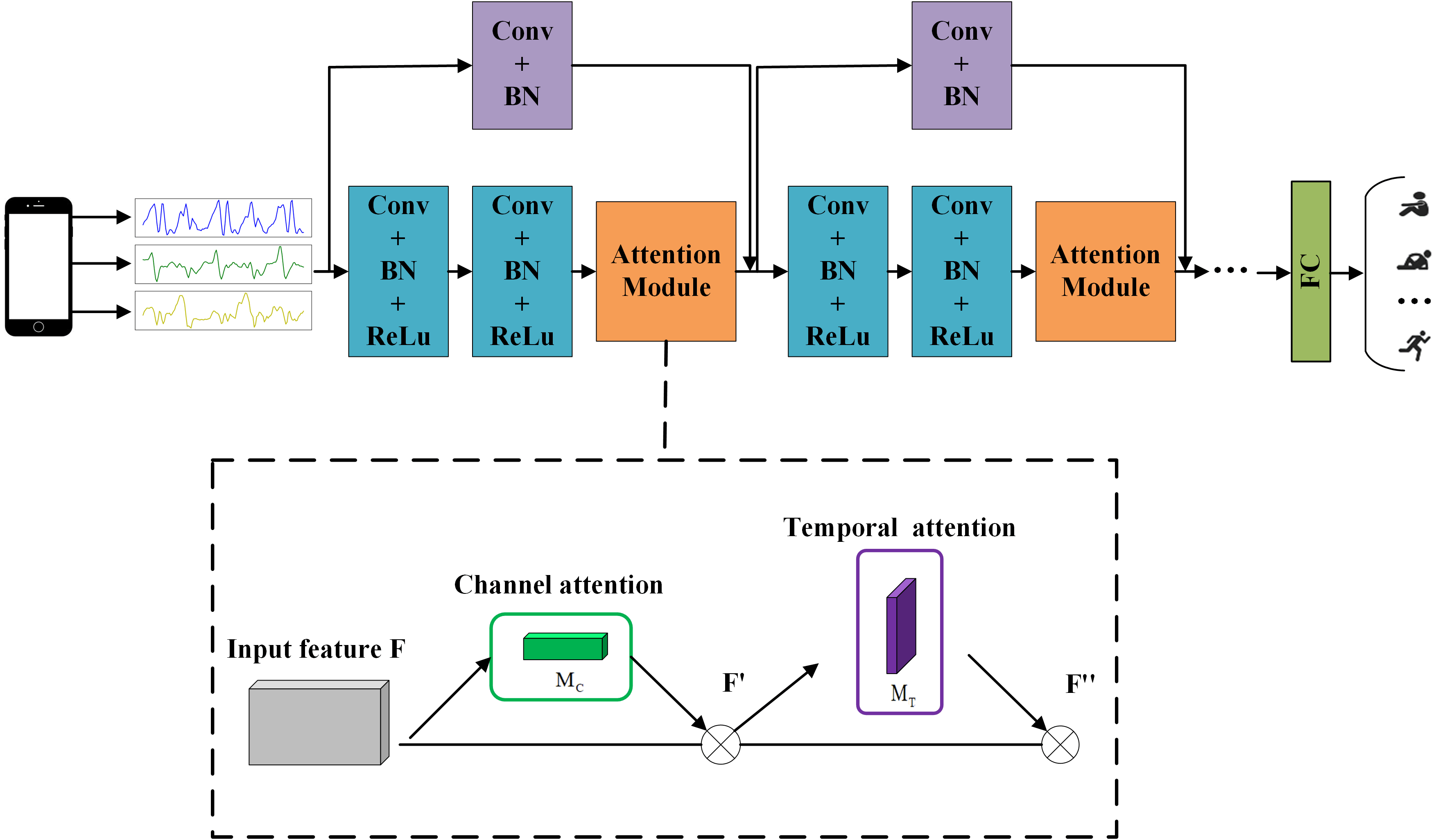}
	\caption  { Overview of the proposed Dual Attention Network.\hspace{2cm}}
\end{figure}
\section{MODEL}
\indent Previous attention framework takes no advantage of sophisticated channel-wise dependencies to mutually promote the learning for both localization and recognition. SENet \cite{hu2018squeeze} for the first time proposes an effective mechanism to learn channel attention and achieves promising performance. Inspired by the recent success of channel attention \cite{hu2018squeeze,gao2019global,woo2018cbam,roy2018recalibrating,cao2019gcnet,fu2019dual}, in this paper, we propose a novel DanHAR that simultaneously considers channel and temporal attention for multimodal time series sensor signals. In addition, the network can be trained end-to-end. For multimodal HAR, one has to firstly deal with multiple channels of time series sensor signals, in which the unique challenges include (i) applying convolutional kernels in CNN along temporal dimension due to various sensor modalities and (ii) then fusing or unifying the features in CNN among multiple sensors. As shown in Fig.1, we design the DanHAR network with convolution, channel attention submodule and temporal attention submodule. First, a fixed-length sliding window is moved along temporal axis to segment time series data into a collection of short pieces of signals, in which an overlap between adjacent windows is tolerated for preserving the continuity of activities. The whole network takes each time window as input, which is then fed into convolutional layers to extract feature representation. Second, the network proceeds to generate channel attention map via employing both max-pooling and average-pooling to aggregate features along temporal axis, followed by a Sigmoid function to produce probabilities. Third, the subsequent temporal attention map is generated by applying both max-pooling and average-pooling techniques to aggerate channel information of a feature map across channel axis.\\
\subsection{Channel Attention Submodule }
\indent Let us consider a CNN layer and its corresponding feature map $A\in R^{C\times H\times W}$, where $C$, $W$ and $H$ are channel (i.e., number of filters), width (temporal axes) and height (sensor axes) dimension. To compute the channel attention efficiently, for a given input feature, SE block \cite{hu2018squeeze} firstly employs global average-pooling to squeeze the temporal dimension for each channel independently, then two fully-connected (FC) layers with non-linearity followed by a Sigmoid function are used to generate channel weights. Recent researches indicate that max-pooling can gather another important clue about distinctive features to infer finer channel-wise attention \cite{hu2018squeeze,woo2018cbam,zhou2016learning}. For more effective feature aggregation, both average-pooled and max-pooled features are used simultaneously. According to SE \cite{hu2018squeeze}, scSE \cite{roy2018recalibrating} and CBAM \cite{woo2018cbam} blocks, the weights of the channel attention can be computed as:
\begin{equation}
\ W_C=\sigma \left( f_{\left\{ w_1,w_2 \right\}}\left( g_1\left( A \right) \right) +f_{\left\{ w_1,w_2 \right\}}\left( g_2\left( A \right) \right) \right) 
\end{equation}
where $g_1=\frac{1}{WH}\sum_{i=1,j=1}^{W,H}{A_{ij}}$ and $g_2=\max _{i=1,j=1}^{W,H}A_{ij}$ are channel-wise global average-pooling and max-pooling. $\sigma$ is a Sigmoid function. As the two FC layers are designed to capture non-linear cross-channel interaction, the Eq.(1) can further be expressed as:
\begin{equation}
W_C=\sigma\left( w_2{ReLU}\left( w_1g_1\left( A \right) \right) +w_2{ReLU}\left( w_1g_2\left( A \right) \right) \right) 
\end{equation}
\noindent in which $ReLU$ indicates the Rectified Linear Unit used between the two FC layers. To reduce parameters, the sizes of $w_1$ and $w_2$ are set to $C\times \left( C/r \right)$ and $\left( C/r \right) \times C$ respectively where $r$ is reduction ratio. Here Eq.(2) make indirect correspondence between channel and its weights, via first projecting channel features into a low-dimensional space and then mapping them back.\\
\subsection{Temporal Attention Submodule}
\indent In a similar way, the temporal attention can be computed by applying average-pooling and max-pooling operations to aggregate channel information of a feature map. There are four different aggregation strategies for the concurrent temporal and channel attention, i.e., max-out, multiplication, addition and concatenation. The addition strategy is used to infer channel attention maps, which provides equal importance to the two pooled features. Referring to the scSE \cite{roy2018recalibrating} and CBAM \cite{woo2018cbam} block, here we concatenate two pooled features to aggregate information along the channel index. The concatenated pooled features are then convolved by a standard convolution layer, which can be formulated as:
\begin{equation}
W_T=\sigma \left( f^{n\times 1}\left( \left[ g_1\left( A \right) ; g_2\left( A \right) \right] \right) \right) 
\end{equation}
where $\sigma$ is a Sigmoid function. $n\times 1$ represents the convolution filter size and $g_1$ , $g_2$ represent average-pooling and max-pooling. The filter size is tuned and set to $7\times 1$. \\
\indent As indicated above, due to multiple sensor modalities, convolution would be applied along temporal dimension and then the features among multiple sensors are unified. Compared to addition strategy, the advantage of this concatenation aggregation lies in that no information is lost. Finally, these two attention submodules are combined via multiplying channel and temporal attention element-wise, each of which is then scaled to [0,1].
\subsection{Residual Network}
\indent Recently, He \emph{et~al.} \cite{he2016deep} proposed to use residual blocks with a simple identity skip-connection to ease the optimization issues of deep networks. Wang \emph{et~al.} \cite{wang2017residual} proposed residual attention network, which incorporates attention mechanism to improve the performance of ResNET on image classification tasks. In the paper, we further utilize residual network as our backbone for the enhancement of feature representing. Up to now, most studies on DNN based HAR still depend on benchmark datasets from conventional machine learning algorithms for evaluation. Building large-scale DNN is prone to overfitting if there is no enough training data. Therefore, the residual network contains only 3 sets of convolutional layers, each of which includes a building block of 2 convolutional layers with the same kernel size in each building block. Totally, there are 6 convolutional layers, and each building block has a short-cut connection that plays a role in skipping the block for identity mapping and adding residual mapping of the block to become the final underlying mapping. The residual network is implemented with double layer skips that contain ReLU and batch normalization in between.\\


\section{Experiments and Results}
\indent The experiments are divided into three parts. The first experiment is to verify that the proposed DanHAR is capable of learning in supervised condition for popular HAR tasks, which are conducted on four public benchmark datasets consisting of WISDM \cite{kwapisz2011activity} dataset, UNIMIB SHAR \cite{micucci2017unimib} dataset, PAMAP2 \cite{reiss2012introducing} dataset and OPPOTUNITY \cite{chavarriaga2013opportunity} dataset. These datasets are recorded in different contexts by either multiple sensor nodes or smartphones worn on the participants, wherein the OPPORTUNITY and PAMAP2 are multimodal HAR datasets. The second experiment is to evaluate the performance of DanHAR on weakly supervised HAR, which uses the weakly labeled sensor dataset previously collected by He \emph{et~al.} \cite{he2018weakly}. In the third experiment, visualizing analysis to the attention weights is provided, which can be used to evaluate the impact of temporal attention weights. In addition, the impact of different sensor modalities placed on different parts of the human body is explored.\\
\indent The dual attention mechanism can easily be incorporated with state-of-the-art feed forward network architectures in an end-to-end training fashion. In the paper, residual network with 3 building blocks is used as backbone to stack our attention modules. As residual building block has seldom been used in HAR researches, for fair comparison, the dual attention mechanism is also embedded into standard CNN via removing skip connections in residual block. Thus, our backbone networks include standard CNN, as well as residual network. Extensive experiments are conducted to analyze the performance improvement caused by attention part. We compare our DanHAR with state-of-the-art DNN models for HAR on each dataset. In the experiments, the number of epochs is set to 500 and Adam optimization method is used to train our model. Our algorithm is implemented in Python by using the machine learning framework Pytorch \cite{paszke2017automatic}. The experiments are all conducted on a workstation with CPU Intel i7 6850k, 64 GB memory, and a NVIDIA 2080ti GPU with 11GB memory. \\

\subsection{Experiment Results and Performance Comparison}

\subsubsection{$\textbf{The WISDM dataset}${\color{black}} \cite{kwapisz2011activity}   }
\indent The WISDM dataset is a public benchmark HAR dataset provided by the Wireless Sensor Data Mining (WISDM) Lab, which contains 6 data attributes: user, activity, timestamp, x-acceleration, y-acceleration, z-acceleration. 29 volunteer participants were enlisted to perform a specific set of activities. In a supervised condition, these participants with an Android smartphone placed in their front pants leg pocket were asked to walk, jog, ascend stairs, descend stairs, sit, and stand for specific periods of time. The data collection was controlled by an application with a simple graphical user interface that executed on the phone. An embedded triaxial accelerometer was used to generate time series sensor data, which was collected every 50ms, i.e., 20 samples per second.\\
\indent Classification algorithms cannot be directly applied to raw time series accelerometer data. Thus, the sliding window technique is firstly applied to transform the raw time series data into examples. To accomplish this, we divided the data into 10-second segments, which are the optimal values used in previous literatures \cite{kwapisz2011activity}. A 95$\%$ overlapping rate is selected, which equals to 0.5s of sliding step length. The whole dataset is randomly split into 70$\%$ training set and 30$\%$ test set respectively. CNN and residual network are used as baselines. When residual connections are removed, the shorthand description of standard CNN architecture is conv128-conv128-conv256-conv256-conv384-conv384-fc, which is comprised of 6 convolutional layers and 1 FC layer. Adam optimizer is used to train with batch size of 210. The initial learning rate is set to 0.001, which will be reduced by a factor of 0.1 after each 50 epochs.

\begin{figure}[htbp]
	\hspace*{0cm}\\ 
	\vspace{-0.8cm}   
	\centering
	\includegraphics[width=9.5cm,height=7cm]{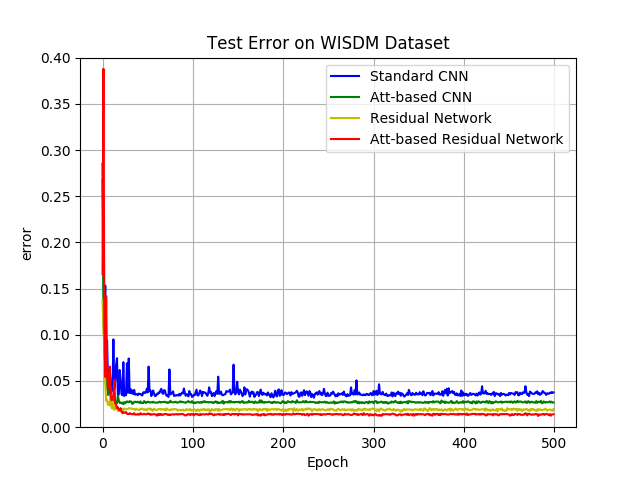}
	\caption{Test errors on \textbf{WISDM} dataset.}
\end{figure}

\indent We employ CNN and residual network as backbone models, and train them with our dual attention submodule. Looking at the test curves (Fig.2) shows that DanHAR consistently outperforms both baselines. The results suggest that dual attention mechanism can be applied to different network structures. We compare DanHAR to baselines, as well as other state-of-the-art DNN models from those literatures (Ignatov \emph{et~al.}, 2018 \cite{ignatov2018real}), (Ravi \emph{et~al.}, 2016 \cite{ravi2016deep}), and (Alsheikh \emph{et~al.}, 2016 \cite{alsheikh2016deep}) accordingly. The results are shown in Table I. It can be seen that our DanHAR performs the best among all the algorithms, with negligible overhead of parameters. DanHAR achieves 0.73$\%$, 0.53$\%$ accuracy improvement on both baselines respectively. To our knowledge, the best published result on the dataset is 98.23$\%$ (Alsheikh \emph{et~al.}, 2016 \cite{alsheikh2016deep}). Our result with dual attention is best reported, which surpasses the state-of-the-art result with a 0.62$\%$ improvement in accuracy.

\begin{table}[htbp]
	\caption{Performance of Different Model Structure for \textbf{WISDM} Dataset.}
	\centering
	\begin{tabular}{ccccc}
		\toprule 
		\textbf{Model}&\textbf{Test Acc}&\textbf{Params}\\
		\midrule
		Standard CNN& 96.83$\%$&1.55M\\
		Att-based CNN& 97.56$\%$&1.57M\\
		Residual Network& 98.32$\%$&2.30M\\
		Att-based Residual Network&\textbf{98.85$\%$}&2.33M\\
		\midrule
		Ignatov \emph{et~al.} 2018 \cite{ignatov2018real}  &93.32$\%$& -\\
		Ravi \emph{et~al.} 2016 \cite{ravi2016deep} &98.20$\%$& -\\
		Alsheikh \emph{et~al.} 2016 \cite{alsheikh2016deep}   &98.23$\%$& -\\
		\bottomrule
	\end{tabular}
\end{table}

\subsubsection{$\textbf{The UNIMIB SHAR dataset}${\color{black}} \cite{micucci2017unimib}}	
\indent UniMiB SHAR is a new smartphone accelerometer dataset designed for benchmarking four different classification tasks: fall vs. no fall, 9 activities, 8 falls, 17 activities and falls. The dataset was collected from 30 subjects wearing a Samsung Galaxy Nexus I9250 smartphone, whose ages range from 18 to 60 years. For each activity, there are 2 to 6 trials performed by each subject. All the activities have 2 trials in which the first one has been recorded with the smartphone placed in left pocket while the other has been recorded with the smartphone in right pocket. On the whole, there are 11,771 samples in the dataset, which are further divided into 17 fine grained classes grouped in two coarse grained classes: 9 types of activities of daily living (ADL) and 8 types of falls.\\
\indent The signals are recorded at a constant sampling frequency of 50 Hz using an embedded triaxial Bosh BMA220 accelerometer. For the labeled accelerometer readings, each signal window of 3s is centered around each peak. That is to say, the dataset is mostly focused on motion-related recognition of ADLs and falls. Thus, instead of overlapped sliding, segmentation technique is directly used. A signal window of 3s is selected, which equals to a signal window with a fixed length T=151. Each accelerometer example is comprised of 3 vectors of 151 values, i.e., a vector of size $1\times 453$. CNN and residual network are used as baselines. With the removal of skip connections, the shorthand description of standard CNN is conv128-conv128-conv256-conv256-conv384-conv384-fc. Adam optimization is used with the batch size of 128. The initial learning rate is set to 1e-3.

\begin{figure}[htbp]
	\hspace*{0cm}\\
	\vspace{-0.8cm}
	\centering
	\includegraphics[width=9.5cm,height=7cm]{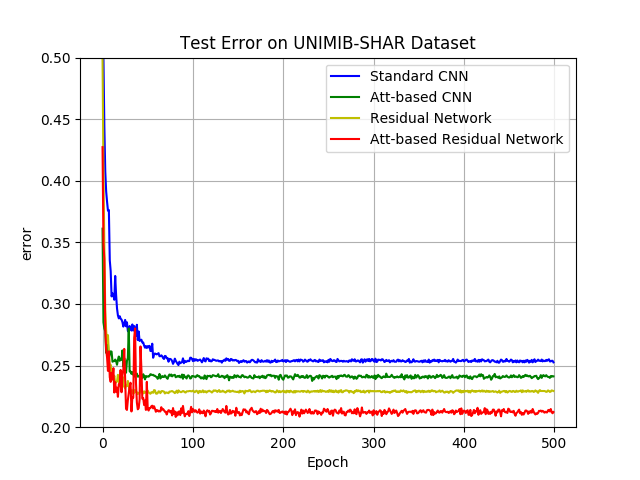}
	\caption{ Test errors on \textbf{UNIMIB SHAR} dataset.}
\end{figure}

\indent In Fig.3, the effectiveness of dual attention mechanism is evaluated in terms of test error curves. It can be clearly seen that the result of our DanHAR is superior to both baselines. Deep models with attention consistently outperform those without attention. Our DanHAR is compared with both baselines and two classical DNN models on the UniMiB SHAR dataset. Results are shown in Table II. As a reference, the two baselines achieve 74.83$\%$ and 77.12$\%$ accuracy. DanHAR surpasses both baselines by 1.24$\%$ and 1.91$\%$ respectively. Attention will not cause any significant increase on the number of parameters. The best reported result on the dataset to our knowledge is 78.07$\%$ using CNN with local loss (Teng \emph{et~al.}, 2020 \cite{TengQi}). Compared with CNN using local loss, DanHAR achieves 0.96$\%$ performance gain in terms of accuracy. Our method also surpasses the Long \emph{et~al.}’s \cite{long2019dual} method by a large margin 2.99$\%$, which uses dual residual networks. 
\begin{table}[htbp]
	\caption{Performance of Different Model Structure for \textbf{UNIMIB SHAR} Dataset.}
	\centering
	\begin{tabular}{ccc}
		\toprule 
		\textbf{Model}&\textbf{Test Acc}&\textbf{Params}\\
		\midrule
		Standard CNN& 74.83$\%$&1.62M\\
		Att-based CNN& 76.07$\%$&1.63M\\
		Residual Network& 77.12$\%$&2.38M\\
		Att-based Residual Network&\textbf{79.03$\%$}&2.40M\\
		\midrule
		Li \emph{et~al.} 2018 \cite{li2018comparison}&74.97$\%$&-\\
		Long \emph{et~al.} 2019 \cite{long2019dual}&76.04$\%$&-\\
		Teng \emph{et~al.} 2020 \cite{TengQi}&78.07$\%$&-\\
		\bottomrule
		\label{Tab2}
	\end{tabular}
\end{table}

\subsubsection{$\textbf{The PAMAP2 dataset}${\color{black}} \cite{reiss2012introducing}}	
\indent The PAMAP2 Physical Activity Monitoring dataset is public available at UCI repository, which contains 18 different physical activities (such as walking, cycling, watching TV, playing soccer, etc.). The dataset was collected from 9 subjects who wore 3 wireless Inertial Measurement Units (IMUs): 1 IMU over the wrist on the dominant arm; 1 IMU on the chest; 1 IMU on the dominant side's ankle. Following a protocol, each of the subjects was instructed to perform 12 different activities such as standing, sitting, ascending stairs, descending stairs, rope jumping and running. Furthermore, some of the subjects also performed a few optional activities such as watching TV, car driving, house cleaning and playing soccer.\\
\indent Accelerometer, gyroscope, magnetometer and heart rate data are recorded from the IMUs on the dominant, resulting in 52 dimensions. A heart rate monitor with sampling rate of 9Hz is also placed on each subject. The sampling rate of the IMUs is 100Hz. For comparison, the IMU signals are downsampled from 100Hz to 33.3Hz. A 5.12 second sliding window with 78$\%$ overlap is selected, resulting in around 473k samples. The shorthand description of the standard CNN is conv128-conv128-conv256-conv256-conv384-conv384-fc. Referring to previous work \cite{hammerla2016deep}, we use subject 5 and 6 for test, and the rest of the subjects for training. Adam optimization is used to train our model. The learning rate is set to 5e-4 and the input batch size is 300. 
\begin{figure}[htbp]
	\hspace*{0cm}\\
	\vspace{-0.9cm}
	\centering
	\includegraphics[width=9.5cm,height=7cm]{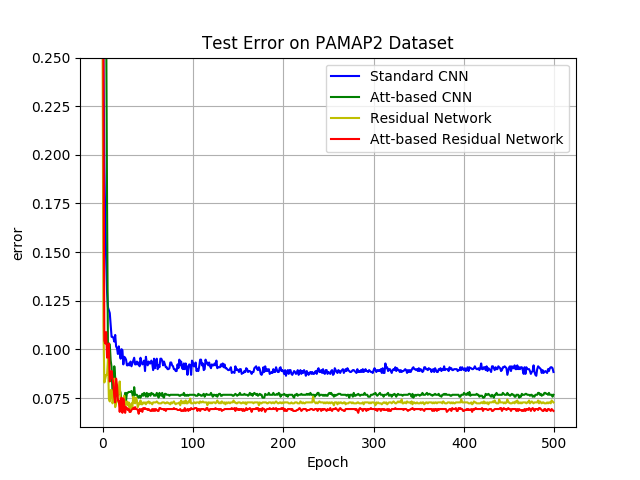}
	\caption{Test errors on \textbf{Pamap2} dataset.}
\end{figure}

\indent  The experiment is conducted on the PAMAP2 dataset to evaluate the performance of DanHAR. CNN and residual network are used as baselines. For each baseline, we compare test error curves when channel attention and temporal attention are superimposed. Fig.4 shows that DanHAR clearly works better than the two tested baseline architectures, achieving lower test errors. We compare our DanHAR with both baselines and state-of-the-art algorithms. Table III shows the results. Our attention method achieves 1.24$\%$ and 0.44$\%$ relative accuracy gain with negligible overhead of parameters, compared with both baselines. The best published result on the dataset to our knowledge is 92.97$\%$ using CNN with local loss (Teng \emph{et~al.}, 2020 \cite{TengQi}). Our result with dual attention is best reported, which surpasses the state-of-the-art results. When compared to the result obtained by Zeng \emph{et~al.} \cite{zeng2018understanding} using attention-based LSTM, our method achieves 3.2$\%$ improvement. Our attention also surpasses AttnSense by a large margin with 3.86$\%$ improvement in accuracy. Comparing with these attention-based recurrent architectures, our DanHAR achieves better performance due to better feature extraction capacity of CNN.

\begin{table}[htbp]
	\caption{Performance of Different Model Structure for \textbf{PAMAP2} Dataset.}
	\centering
	\begin{tabular}{ccc}
		\toprule 
		\textbf{Model}&\textbf{Test Acc}&\textbf{Params}\\
		\midrule
		Standard CNN& 91.21$\%$&2.73M\\
		Att-based CNN& 92.45$\%$&2.75M\\
		Residual Network& 92.72$\%$&3.48M\\
		Att-based Residual Network&\textbf{93.16$\%$}&3.51M\\
		\midrule
		Zeng \emph{et~al.} 2018  \cite{zeng2018understanding}    &89.96$\%$&-\\
		Ma   \emph{et~al.} 2018  \cite{ma2019attnsense}    &89.30$\%$&-\\
		Teng \emph{et~al.} 2020 \cite{TengQi}&92.97$\%$&-\\
		\bottomrule
		\label{Tab2 }
	\end{tabular}
\end{table}

\subsubsection{$\textbf{The OPPORTUNITY dataset}${\color{black}} \cite{chavarriaga2013opportunity}}	
\indent The OPPORTU-NITY dataset records realistic daily life activities of 12 subjects in a sensor-rich environment. Overall, in a daily living scenario, 15 networked sensor systems with 72 sensors of 10 modalities, were integrated in the environment and on the body. In the paper, we employ the same subset in previous OPPORTUNITY challenge, which consists of 18 annotated mid-level gesture recordings from on-body sensors from 4 subjects. Each subject was instructed to perform five different runs of common kitchen activities. Data was recorded at a sampling frequency of 30Hz. Sensor recordings include IMUs attached at 12 on-body position, i.e., the upper limbs, the back, and both feet. The resulting dataset has 79 dimensions.\\ 
\indent As the OPPORTUNITY dataset has a notable imbalanced class distribution with an additional strong bias towards the NULL class that represents 72.28$\%$, model performance is evaluated considering the NULL class. The same training and test protocol are replicated from (Hammerla \emph{et~al.}, 2016 \cite{hammerla2016deep}). The 2, 4 and 5 run from subject 1, 2 and 3 are used as test set, and the remaining data are used for training. Residual network is used as backbone to construct our dual attention network. When skip connections are removed, the shorthand description of standard CNN is conv128-conv128-conv256-conv256-conv384-conv384-fc. For frame-by-frame analysis, we perform classification in a sliding window of 64, with a step length of 8. The batch size is set to 300. The initial learning rate is set as 0.0001, which will be reduced by a factor of 0.1 after each 50 epochs. Adam optimization is used with default parameters in Pytorch \cite{paszke2017automatic}.\\ 

\begin{figure}[htbp]
	\hspace*{0cm}\\
	\vspace{-1.2cm}
	\centering
	\includegraphics[width=9.5cm,height=7cm]{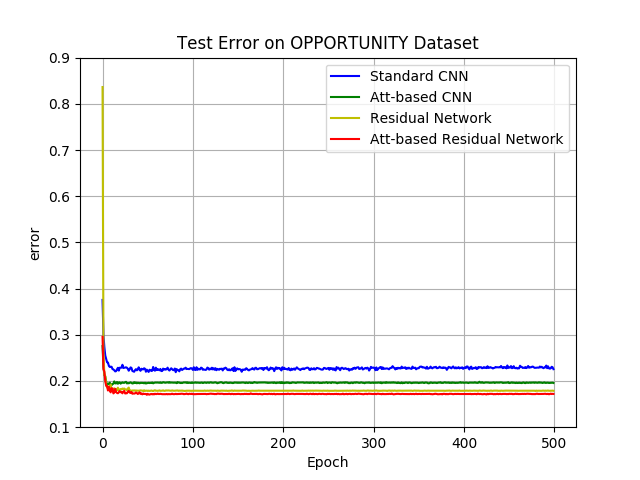}
	\caption{  Test errors on \textbf{OPPORTUNITY} dataset.}
\end{figure}

\indent Test error curves of DanHAR and both baselines on the OPPORTUNITY dataset are presented in Fig.5. It can be clearly seen that DanHAR generally achieves lower test errors than the corresponding baselines. We compare DanHAR with both baselines, as well as other state-of-the-art methods. Results are shown in Table IV, which includes a comprehensive list of past published deep learning techniques from those literatures (Zeng \emph{et~al.}, 2014 \cite{zeng2014convolutional}), (Ordóñez \emph{et~al.}, 2016 \cite{ordonez2016deep}), (Teng \emph{et~al.}, 2020 \cite{TengQi}), and (Hammerla \emph{et~al.} 2016 \cite{hammerla2016deep}). The results show that DanHAR achieve 2.89$\%$ and 0.7$\%$ performance improvement over standard CNN and residual network respectively. It can be seen that DanHAR performs the best among all the algorithms. The best published result on this task is to our knowledge 81$\%$ using CNN with local loss (Teng \emph{et~al.}, 2020 \cite{TengQi}). Our DanHAR achieves around 1.75$\%$ improvements in terms of accuracy. The second best result is 78.90$\%$ using DeepConvLSTM. DanHAR outperforms DeepConvLSTM by a large margin with 3.85$\%$ improvement in accuracy.

\begin{table}[htbp]
	\caption{Performance of Different Model Structure for \textbf{OPPORTUNITY} Dataset.}
	\centering
	\begin{tabular}{ccc}
		\toprule 
		\textbf{Model}&\textbf{Test Acc}&\textbf{Params}\\
		\midrule
		Standard CNN& 77.53$\%$&1.15M\\
		Att-based CNN& 80.42$\%$&1.17M\\
		Residual Network& 82.05$\%$&1.55M\\
		Att-based Residual Network&\textbf{82.75$\%$}&1.57M\\
		
		\midrule
		Zeng \emph{et~al.} 2014 \cite{zeng2014convolutional}&76.83$\%$&-\\
		Ordóñez   \emph{et~al.} 2016 \cite{ordonez2016deep}   &78.90$\%$&-\\		
		Hammerla   \emph{et~al.} 2016 \cite{hammerla2016deep}   &74.50$\%$&-\\
		Teng \emph{et~al.} 2020 \cite{TengQi}&81.00$\%$&-\\
		\bottomrule
	\end{tabular}
\end{table}

\begin{figure}[htbp]
	\hspace*{0cm}\\
	\vspace{0cm}
	\centering
	\includegraphics[width=8cm,height=2.5cm]{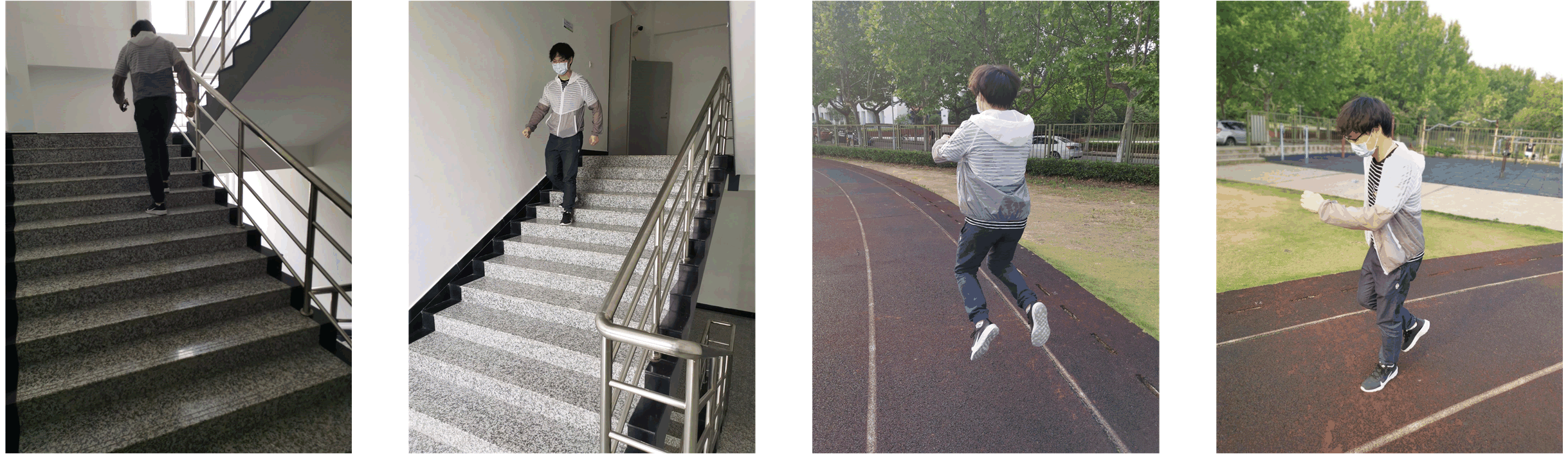}
	\caption{Data collection process. The four images from left to right are going upstairs, going downstairs, jumping and jogging.}
\end{figure} 

\subsubsection{$\textbf{The Weakly Labeled dataset}${\color{black}}}	
\indent Our weakly labeled dataset was collected using a triaxial acceleration sensor embedded in iPhone 7. During the experiment, 10 participants were asked to place the smartphone in their right trouser pocket to perform human daily activities. Fig.6 shows the data collection process for one participant. The dataset contains five kinds of activities: walking, jogging, jumping, going upstairs and going downstairs, wherein walking is seen as the background activity, while the other four activities are the target activities to be recognized. Each participant performs 4 runs for one specific activity. Data collection was made through an application called HascLogger, which can collect activity data by iPhone in real time. The user interface of the application is shown in Fig.7.\\
\begin{figure}[htbp]
	\hspace*{0cm}\\
	\vspace{0cm}
	\centering
	\includegraphics[width=8cm,height=4cm]{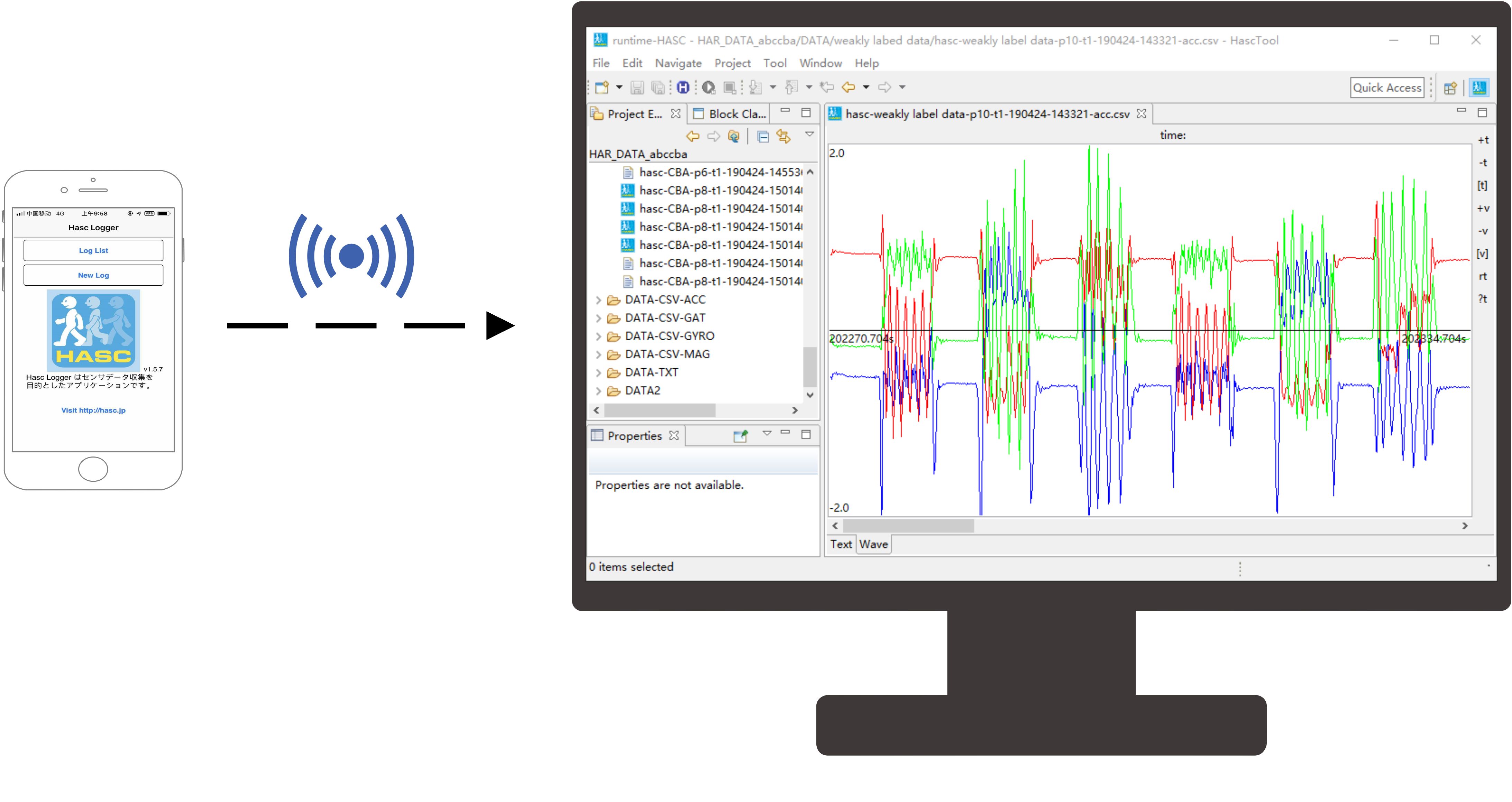}
	\caption{The user interface of the acquisition software HascLogger.}
\end{figure}

\begin{table}[htbp]
	\caption{Weakly Labeled Dataset Statistics.}
	\centering
	\begin{tabular}{ccc}
		\toprule 
		\textbf{Activity}&\textbf{Label}&\textbf{Number}\\
		\midrule
		going upstairs& 0 &20194\\
		going downstairs& 1 &18611\\
		jumping& 2 &14088\\
		jogging& 3 &23264\\
		\midrule
		total& - &76157\\
		\bottomrule
	\end{tabular}
\end{table}

\indent The data is recorded at a sampling frequency of 50Hz. A sliding window of 40.96 seconds (window width is 2048) is used to segment the time series sensor data, resulting in around 912K examples. The weakly labeled 2048-length data samples may contain one interesting activity or multiple interesting activities. The statistics of different activity samples are presented in Table V. CNN and residual network are used as baselines. The dataset is split into 70$\%$ training set and 30$\%$ test set. Adam optimization is used to train model with mini batch size of 200. The initial learning rate is set to 1e-3.\\ 
\begin{figure}[htbp]
	\hspace*{0cm}\\
	\vspace{-1.2cm}
	\centering
	\includegraphics[width=9.5cm,height=7cm]{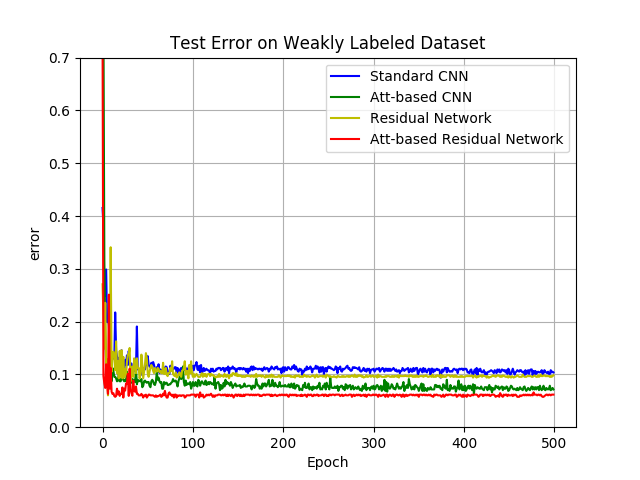}
	\caption{Test errors on the \textbf{Weakly Labeled} dataset.}
\end{figure}

\indent Fig.8 shows that our attention method consistently outperforms the two baselines, achieving lower test error. We compare our DanHAR with both baselines, as well as several state-of-the-art algorithms on the weakly labeled dataset. The results are shown in Table VI. It can be seen that our DanHAR performs the best among all the algorithms, achieving 3.76$\%$ and 3.93$\%$ accuracy improvements over both baselines respectively. Our DanHAR outperforms DeepConvLSTM by a considerable margin of 4.82$\%$. Comparing with our previous work, the attention method achieves 1.03$\%$ improvement on the dataset. The results indicate that the channel-wise attention plays an important role in improving performance for the weakly supervised learning task.\\

\begin{table}[htbp]
	\caption{Performance of Different Model Structure for the \textbf{Weakly Labeled} Dataset.}
	\centering
	\begin{tabular}{ccc}
		\toprule 
		\textbf{Model}&\textbf{Test Acc}&\textbf{Params}\\
		\midrule
		Standard CNN& 89.86$\%$&0.70M\\
		Att-based CNN& 93.62$\%$&0.71M\\
		Residual Network& 90.93$\%$&0.94M\\
		Att-based Residual Network&\textbf{94.86$\%$}&0.95M\\
		
		\midrule
		DeepConvLSTM \cite{wang2019attention}  &90.04$\%$&-\\		
		Attention-based CNN \cite{wang2019attention}  &93.83$\%$&-\\
		\bottomrule
	\end{tabular}
\end{table}

\subsection{The Ablation studies }
\indent Through plugging the dual attention module, we obtain accuracy improvement from the two backbone networks across various HAR datasets, which implies that the performance boost may come from accurate information, e.g., what and where to attention. We further perform several ablation experiments to thoroughly evaluate the improvements for interpretability of models. The experiments are conducted on the WISDM dataset, the PAMAP2 dataset and the weakly labeled dataset, and all hyper-parameters are exactly the same as used above.\\
\begin{figure}[htbp]
	\hspace*{0cm}\\
	\vspace{-1.2cm}
	\centering
	\includegraphics[width=9.5cm,height=7cm]{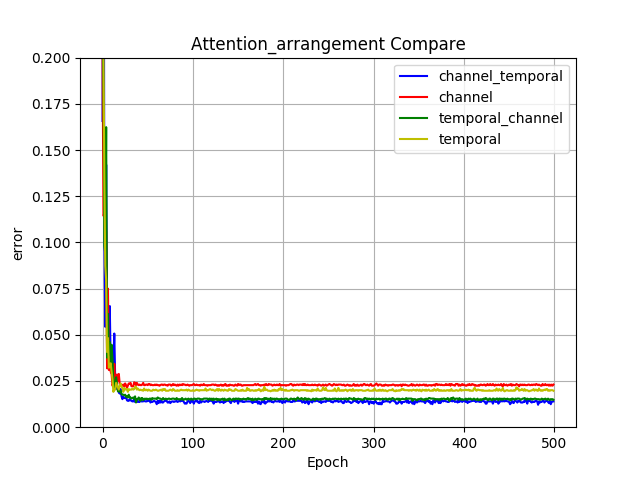}
	\caption{Effects of single attention mechanism and superimposed attention mechanism on \textbf{WISDM} dataset.}
\end{figure}

\indent First, we study the influence of arranging the channel and temporal attention submodules in different ways: sequential channel-temporal attention, sequential temporal-channel attention, channel attention alone, and temporal attention alone. As each attention module has its unique function, the arrangement may affect final overall performance. For example, the channel attention tends to tell what to attention, while the temporal attention tends to tell where to attention. Fig.9 shows the experimental results in different arranging ways of attention. It can be seen from Table VII that sequential channel-temporal attention performs the best among all the arrangements, achieving 0.23$\%$, 0.73$\%$, 0.76$\%$ performance improvement over sequential temporal-channel attention, channel attention alone and temporal attention alone respectively. The results suggest that dual attention is superior to one attention alone. In addition, the channel-first order is slightly superior to the temporal-first order.\\
\begin{table}[htbp]
	\caption{Performance of Different Arrangement of Attention Modules for \textbf{WISDM} Dataset.}
	\centering
	\begin{tabular}{ccc}
		\toprule 
		\textbf{Module}&\textbf{Test Acc}&\textbf{Params}\\
		\midrule
		Channel& 98.09$\%$&2.32M\\
		Temporal& 98.12$\%$&2.30M\\
		Temporal+Channel& 98.62$\%$&2.32M\\
		Channel+Temporal&\textbf{98.85$\%$}&2.32M\\
		\bottomrule
	\end{tabular}
\end{table}

\indent Second, to better understand how channel and temporal attention achieve accuracy improvement, we provide visualizing analysis which helps to determine which are the important signal components by looking into the channel or temporal attention weights of the models. We visualize the temporal attention weights on the weakly labeled dataset. Fig.10 shows 3 sensor signal windows which are coarsely labeled as jogging, going downstairs and jumping. Unlike strictly labeled benchmark dataset, the entire signal segment contains the target activity, as well as a lot background activity noise (i.e., walking). Our attention method prefers to put a high emphasis on a small portion of the interesting activity and ignore such background activities, which indicates that DanHAR can capture the important part from a long sequence to increase performance, as well as sensor signal’s comprehensibility. The result also implies that DanHAR can aid to the process of sensor data annotation, and make the collection of “ground truth labeled” data easier.\\
\begin{figure}[htbp]
\centering

\subfigure[jogging]{	
\begin{minipage}{0.5\linewidth}
		\centering
		\hspace*{-2.6cm}
		\includegraphics[width=9cm,height=4.5cm]{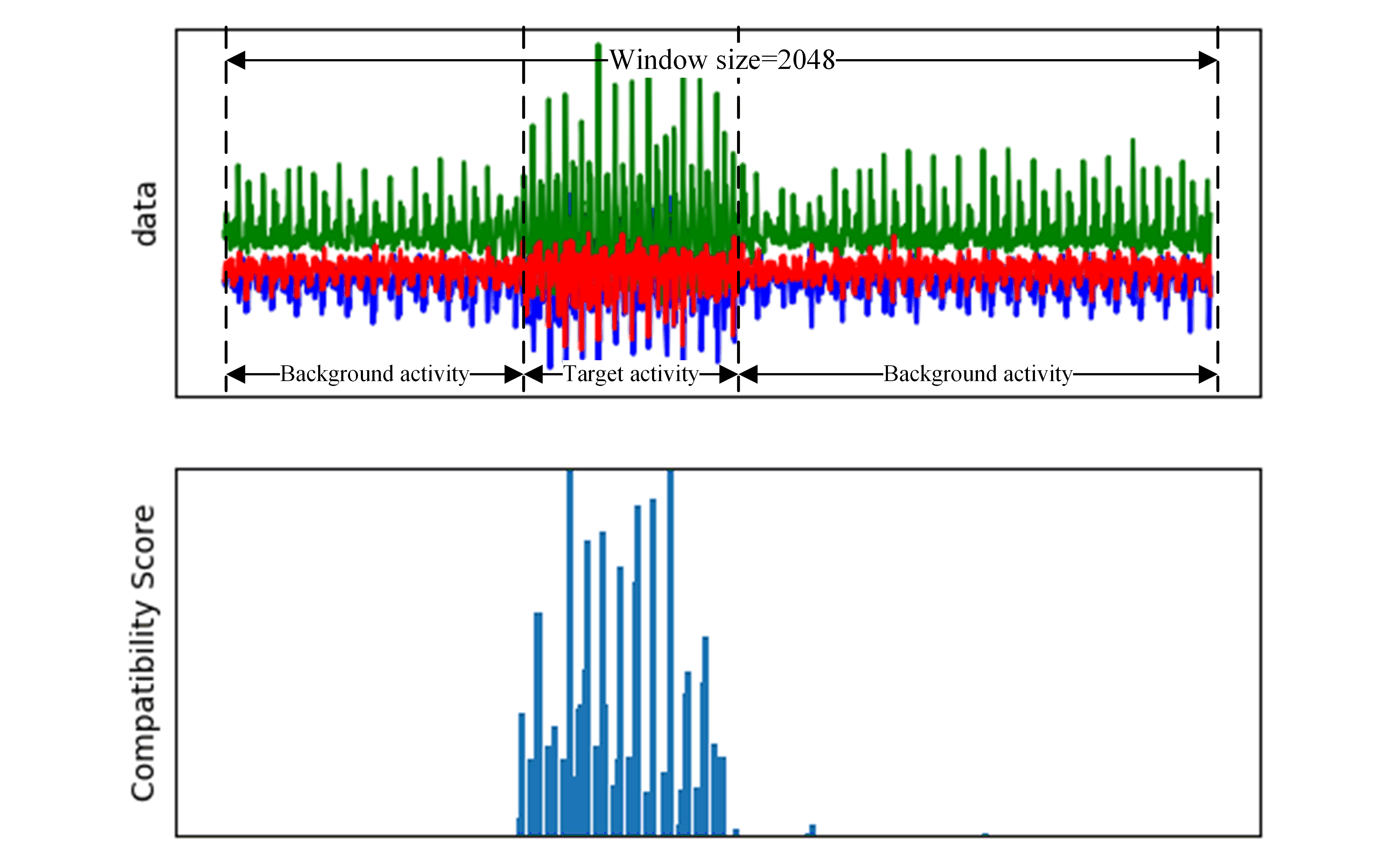} 	
		\centering
\end{minipage}}

\subfigure[going downstairs]{
\begin{minipage}{0.5\linewidth}
		{\centering\hspace*{-2.5cm}\includegraphics[width=9cm,height=4.5cm]{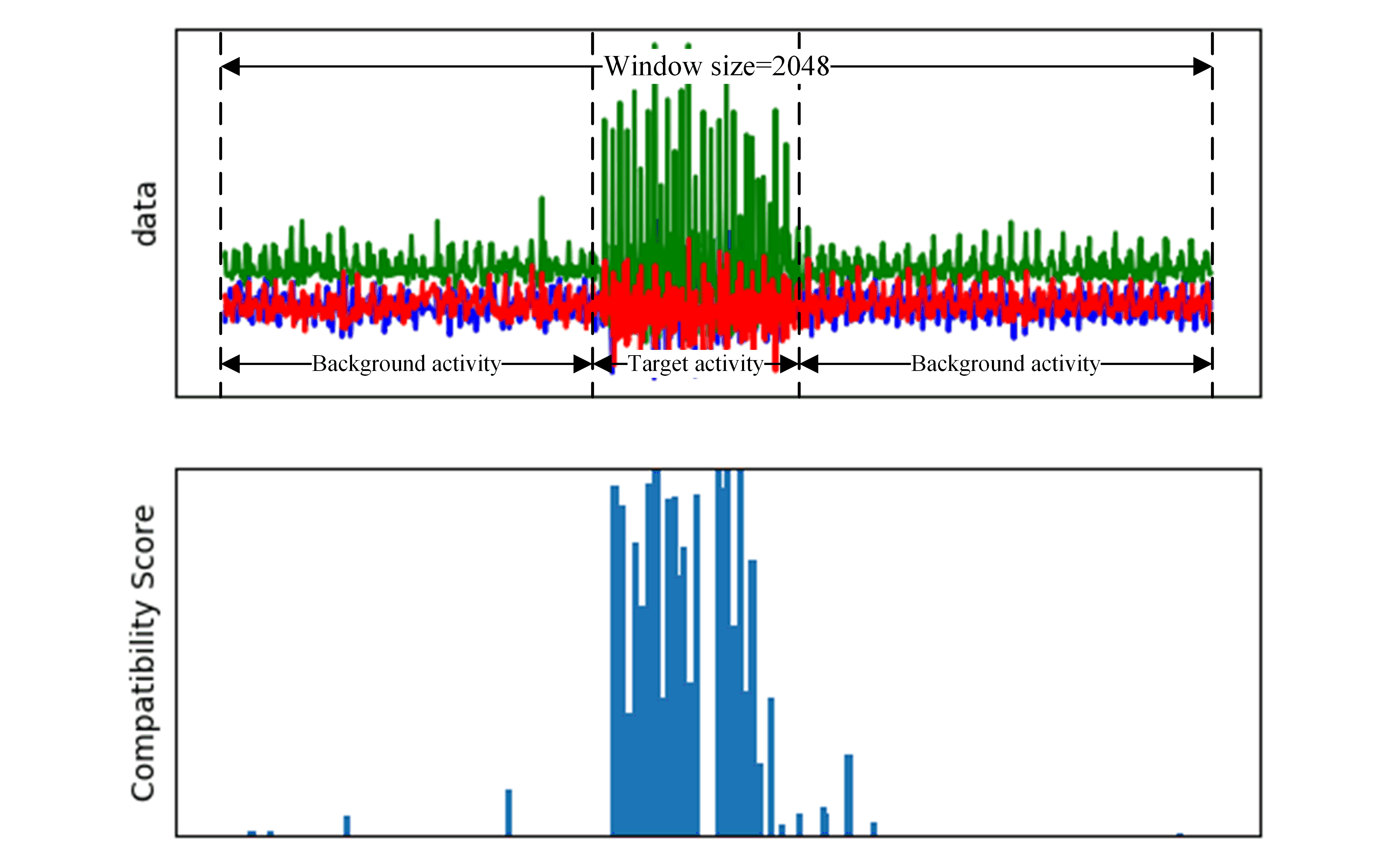}}
		\centering	
\end{minipage}}

\subfigure[jumping]{
\begin{minipage}{0.5\linewidth}
		{\centering\hspace*{-2.5cm}\includegraphics[width=9cm,height=4.5cm]{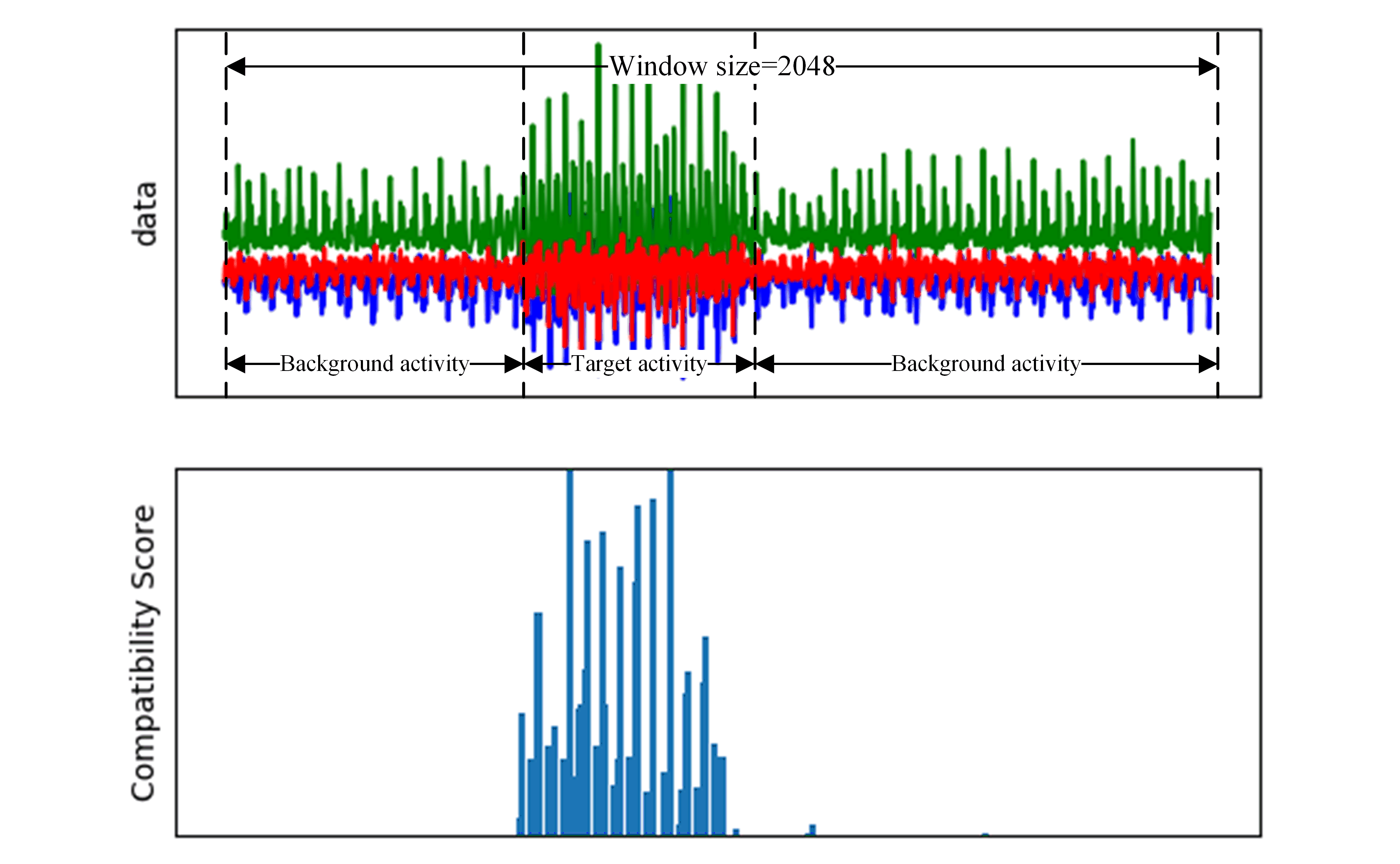}}		 	
		\centering
\end{minipage}}	

	\centering
	\caption{Illustration of the weakly sensor data sample. The attention modules which we proposed can locate the target activity well.}
\end{figure}

\indent Third, we also provide visualizing analysis of the channel attention weights, which can be used to evaluate the impact of different sensor modalities placed on different parts of the human body. As there is only a smartphone used in the weakly labeled dataset, instead we choose to use the multimodal PAMAP2 dataset in the experiment. Fig.11 shows the positions of IMUs placed on human body. Fig.12(a)(b)(c) shows the attention weights of different sensor modalities for the various human activities. Compared with our previous temporal attention mechanism that treats all the sensor modality equally, the dual attention mechanism can automatically learn the priority of different sensors, which performs better in feature fusion and achieves better performance in dealing with multimodal HAR tasks. As can be seen in Fig.12, for running activity, the channel attention puts a high emphasis on the hand sensor (Hand$\_$y, Hand$\_$z), the chest sensor (Chest$\_$x, Chest$\_$y), and the ankle sensor (Ankle$\_$x, Ankle$\_$y). For rope jumping activity, the channel attention put more emphasis on the hand sensor (Hand$\_$x) ,the chest sensor (Chest$\_$z), and the ankle sensor (Ankle$\_$z). For cycling activity, our method pays more attentions on the chest sensor (Chest$\_$z) and the ankle sensor (Ankle$\_$z). The results are interpretable and consistent with our common intuition.\\
\begin{figure}[htbp]
	\hspace*{0cm}\\
	\vspace{0cm}
	\centering
	\includegraphics[width=3cm,height=4cm]{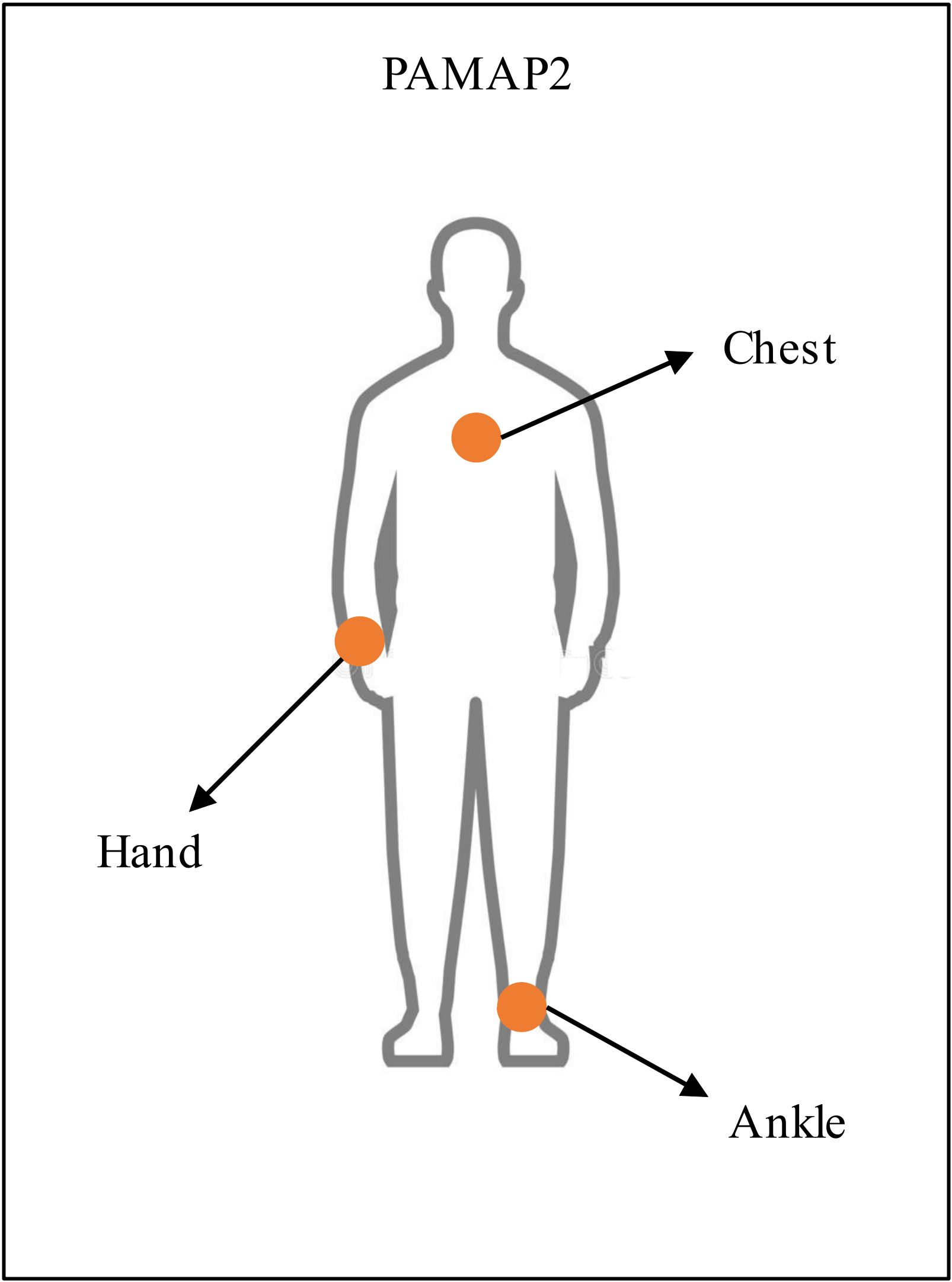}
	\caption{\label{Fig4}   Positions of sensors in PAMAP2 dataset. }
\end{figure}
\begin{figure}[htbp]
	\centering	

	\subfigure[]{
	\begin{minipage}{0.5\linewidth}
		\centering
		\hspace*{-1.4cm}
		\includegraphics[width=7cm,height=5cm]{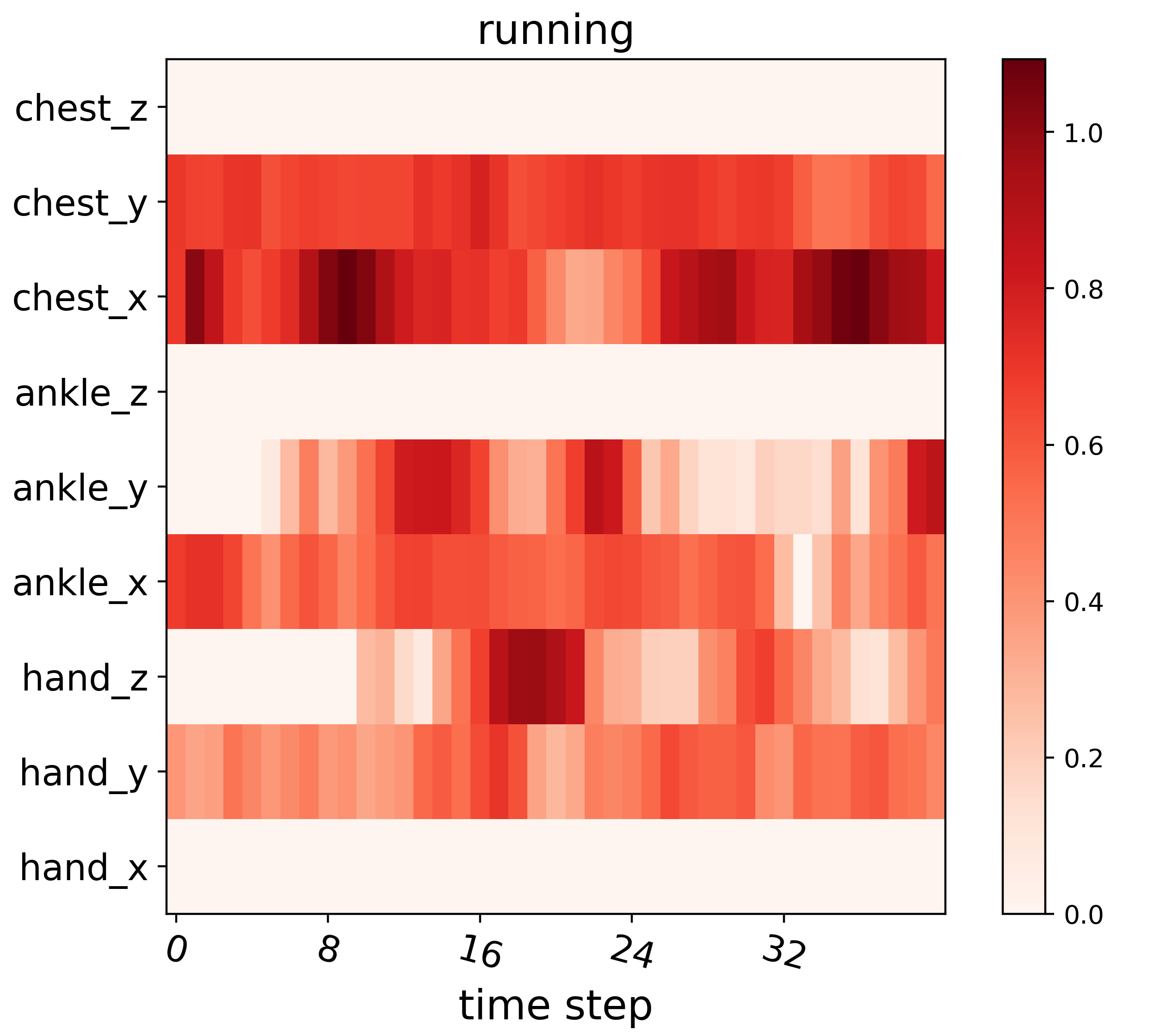}	 	
		\centering
	\end{minipage}}
	
	\subfigure[]{
	\begin{minipage}{0.5\linewidth}
		{\centering\hspace*{-1.4cm}
		\includegraphics[width=7cm,height=5cm]{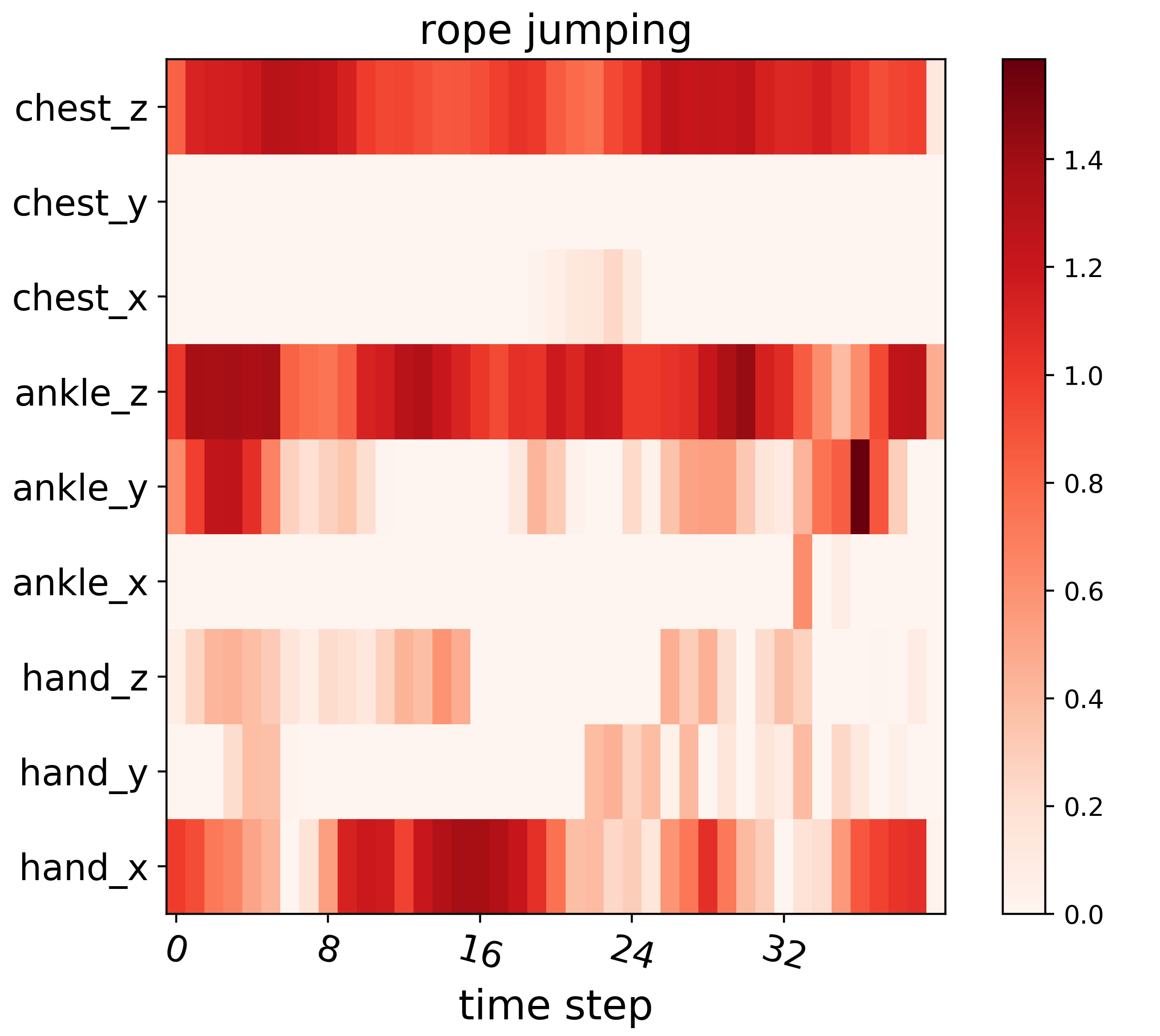}}
		\centering
	\end{minipage}}

	\subfigure[]{
	\begin{minipage}{0.5\linewidth}
		{\centering\hspace*{-1.4cm}
		\includegraphics[width=7cm,height=5cm]{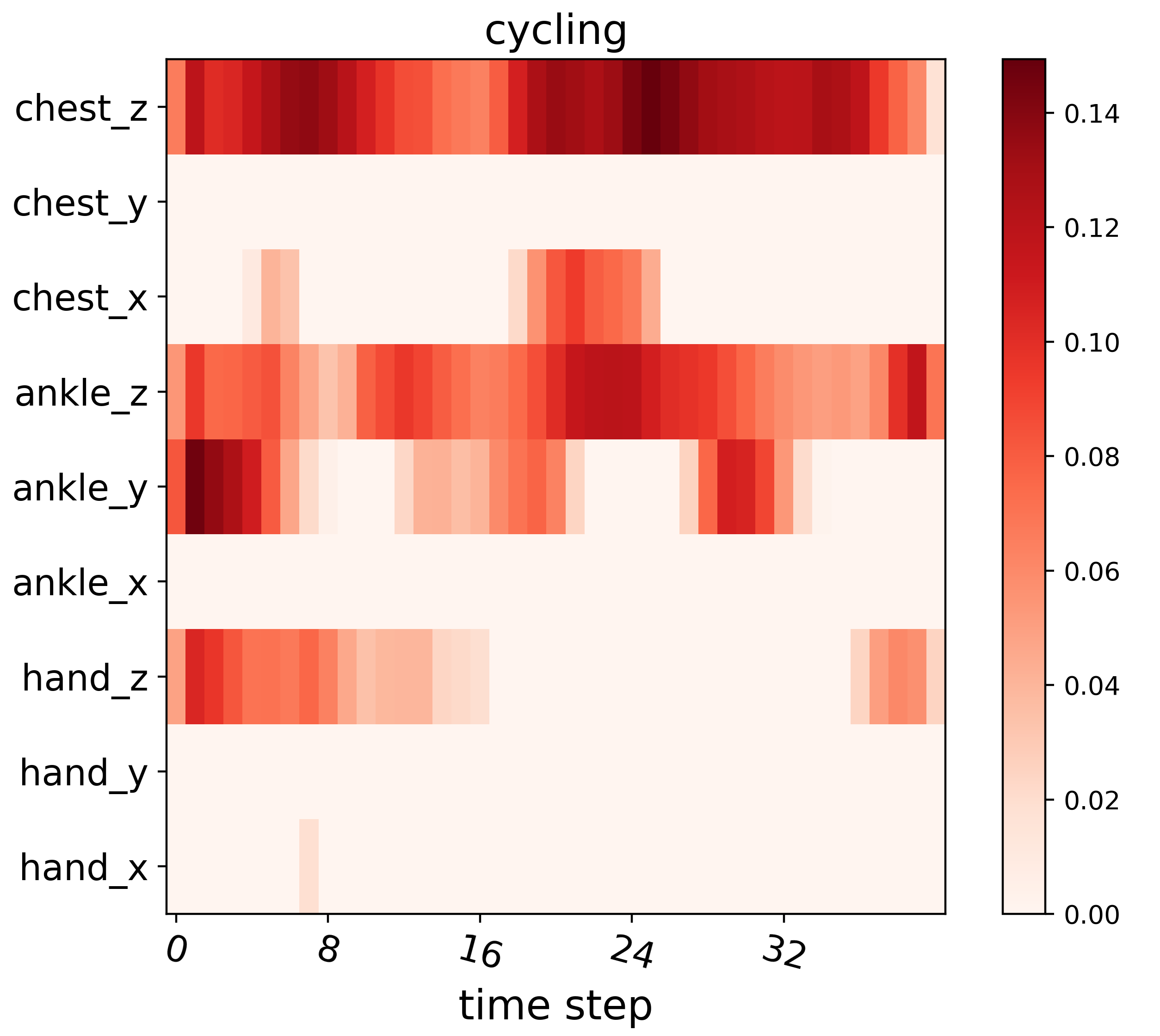}}		 	
		\centering
	\end{minipage}}
	
	\centering
	\caption{Visualization of channel attention weights of various activities in PAMAP2 dataset.}
\end{figure}

\indent Finally, we compare the confusion matrices of DanHAR and residual network on the PAMAP2 dataset. Fig.13 shows that many of the misclassifications are due to confusion between two very similar activity classes like rope jumping and walking, which were previously perceived to be very difficult to distinguish. This may be attributed to their similar vibrations in signal waveforms. From the results, we find that residual network makes 93 errors, while DanHAR misclassifies only 75 activities, which confirms the superiority of the proposed method in recognition accuracy.
\begin{figure}[htbp]
		\begin{minipage}{0.4\linewidth}
		\centering
		\vspace*{0cm}
		\hspace{-0.4cm}
		\includegraphics[width=9.8cm,height=7.5cm]{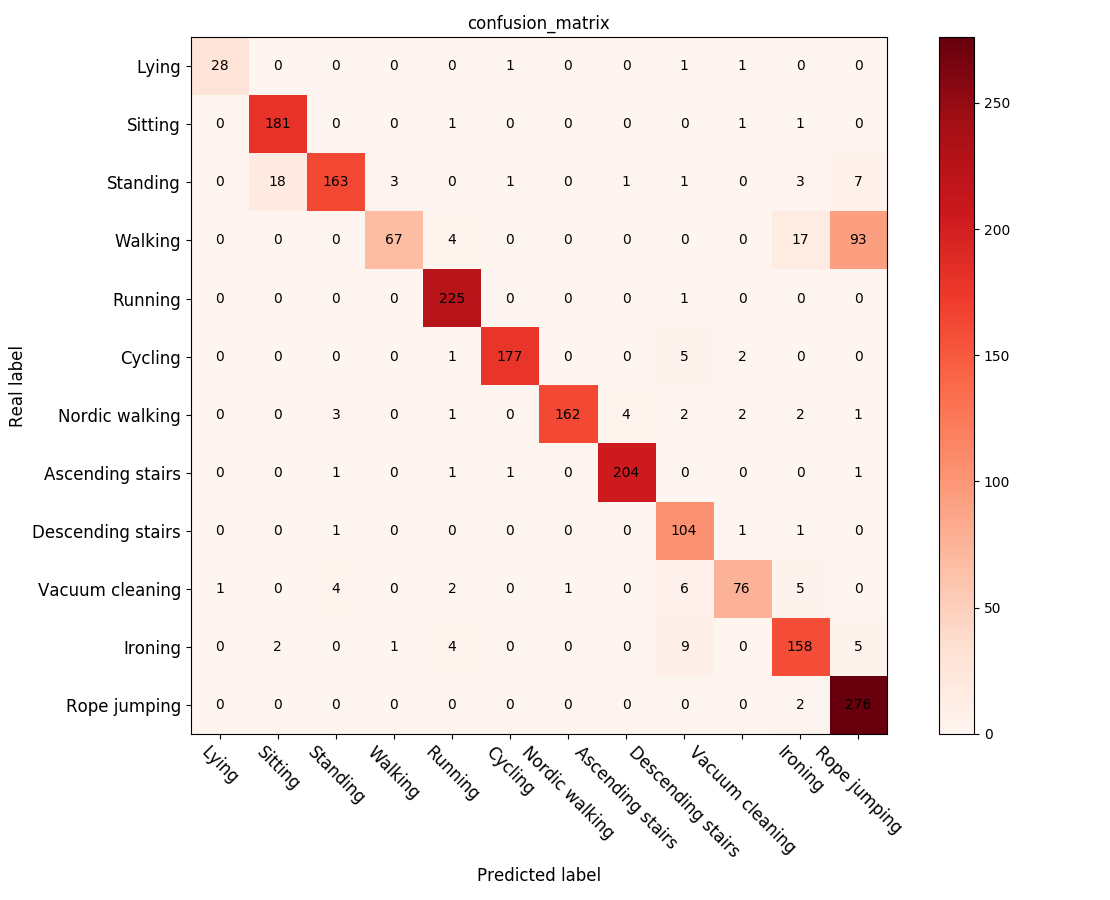}	 	
		\centering\hspace*{0cm}
\end{minipage}

\begin{minipage}{0.4\linewidth}
		\hspace{-0.4cm}
		{\centering\includegraphics[width=9.8cm,height=7.5cm]{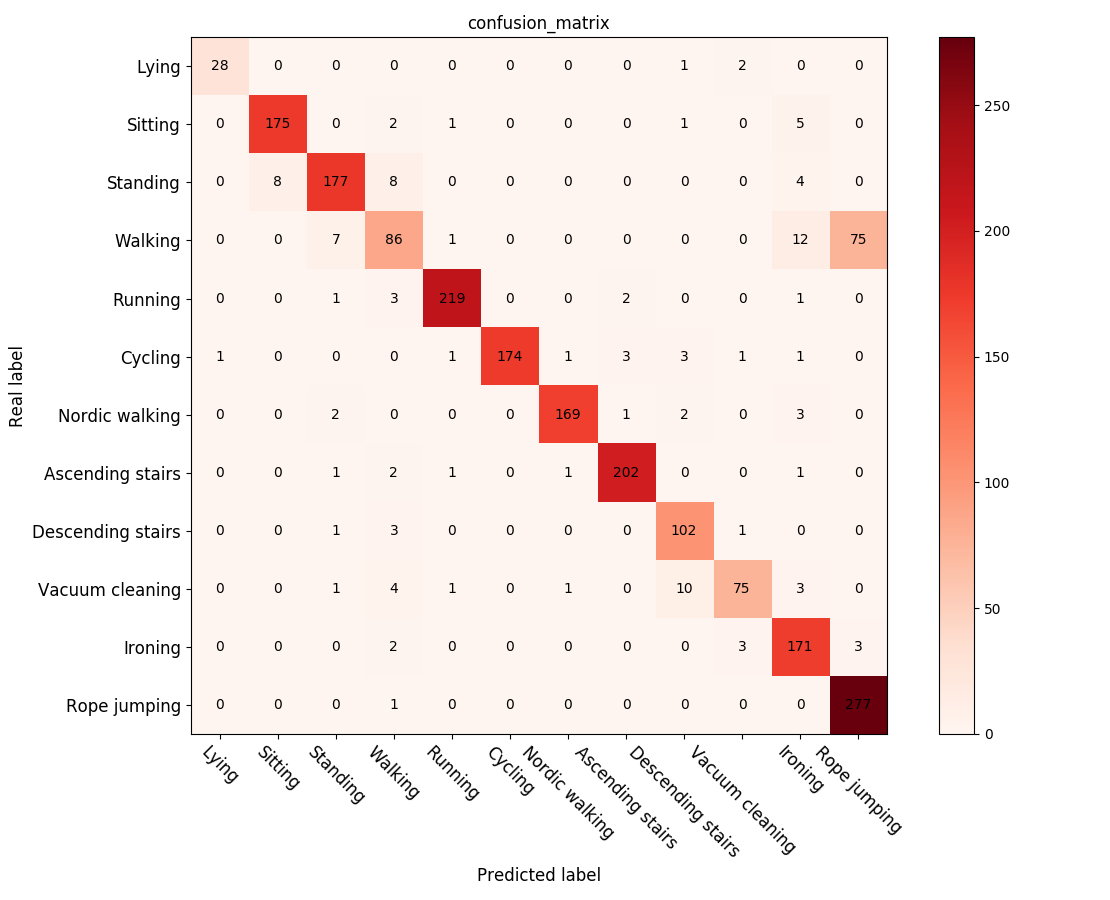}}
		\centering\hspace*{0cm}			
\end{minipage}
\caption{Confusion matrix for \textbf{PAMAP2} dataset between residual network and DanHAR from top to bottom.}
\end{figure}

\section{Conclusion}
\indent As a comparison, He \emph{et~al.} \cite{he2018weakly} recently proposed a hard temporal attention method using reinforcement learning in weakly supervised HAR scenario, which is hard to train in an end-to-end manner. We proposed an end-to-end trainable method, which utilizes only one soft temporal attention \cite{wang2019attention}. However, using temporal attention alone can only tells where to focus, and miss the channel attention, which plays an important role in deciding what to focus. Therefore, the two attention-based methods are only able to determine how much data from each signal window contribute to the classification, which makes them hard to focus on important sensor modalities. That is to say, the two existing methods fail to address the spatial-temporal dependencies of the sensing signals in various multimodal HAR tasks. In another line of research, two previous attention methods combining with LSTM \cite{zeng2018understanding} or GRU \cite{ma2019attnsense} are proposed to capture the dependencies of sensing signals in both spatial and temporal domains, which shows advantages in improving sensor singal’s comprehensibility. However, they heavily rely on recurrent architecture to interpret the recognition process, which has weak feature representing power. \\
\indent Recognizing human activities from multimodal sensor data is a challenging task. In the paper, we for the first time propose dual attention method called DanHAR, which uses channel attention and temporal attention simultaneously to better understand and improve deep networks for various multimodal HAR tasks. DanHAR adopted a hybrid framework to combine dual attention mechanism with CNN or residual network to fuse multimodal sensor information, which has better capability to capture temporal-spacial patterns in multimodal sensing data for HAR. The proposed DanHAR method is evaluated with various state-of-the-art methods on four public HAR datasets consisting of WISDM dataset \cite{kwapisz2011activity}, PAMAP2 dataset \cite{reiss2012introducing}, UNIMIB SHAR dataset \cite{micucci2017unimib}, and OPPORTUNITY dataset \cite{chavarriaga2013opportunity}, as well as the weakly labeled dataset. The results confirm that DanHAR achieves significant performance improvement meanwhile keeping the overhead of parameters small. Various ablation experiments are performed. We visualize how the attention method exactly infers given an input activity. On the whole, the proposed attention mechanism can automatically learn the priority of different sensors and capture the important part from a long sequence to increase sensor signal’s comprehensibility. In the future, we will explore different applications of DanHAR to better understand mixed attention mechanism for specific multimodal HAR task.


%




\bibliographystyle{IEEEtran}
\bibliography{ref}


%




\end{document}